\documentclass[10pt,twocolumn,letterpaper]{article}

\usepackage[pagenumbers]{cvpr} 

\usepackage{graphicx}
\usepackage{amsmath}
\usepackage{amssymb}
\usepackage{booktabs}
\usepackage{bm}
\usepackage{float}

\usepackage{stfloats} 

\usepackage{dsfont}
\usepackage{subcaption}
\usepackage{caption}
\usepackage{nicefrac}
\usepackage{mathrsfs}
\usepackage{xcolor}
\usepackage{enumitem}
\usepackage{colortbl}
\usepackage{multirow}
\usepackage[pagebackref=true,breaklinks=true,colorlinks,bookmarks=false,citecolor=blue,linkcolor=blue,urlcolor=blue]{hyperref}

\usepackage{marvosym}
\usepackage[export]{adjustbox}

\usepackage{pifont}

\def\xnet{BIPNet\xspace}
\usepackage[capitalize]{cleveref}
\crefname{section}{Sec.}{Secs.}
\Crefname{section}{Section}{Sections}
\Crefname{table}{Table}{Tables}
\crefname{table}{Tab.}{Tabs.}

\def\cvprPaperID{8335} 
\def\confName{CVPR}
\def\confYear{2022}

\usepackage{capt-of,etoolbox}

\begin{document}

\title{Burst Image Restoration and Enhancement}

\author{Akshay Dudhane$^1$ \quad Syed Waqas Zamir$^2$ \quad Salman Khan$^{1,3}$ \quad \\
Fahad Shahbaz Khan$^{1,4}$ \quad Ming-Hsuan Yang$^{5,6,7}$ \\
$^1$Mohamed bin Zayed University of AI \hspace{1.5mm} $^2$Inception Institute of AI \hspace{1.5mm} $^3$Australian National University\\
 $^4$Link\"{o}ping University \hspace{1.5mm} $^5$University of California, Merced \hspace{1.5mm} $^6$Yonsei University \hspace{1.5mm} $^7$Google Research}

\maketitle

  
\begin{abstract}
    Modern handheld devices can acquire burst image sequence in a quick succession. However, the individual acquired frames suffer from multiple degradations and are misaligned due to camera shake and object motions. 
    The goal of Burst Image Restoration is to effectively combine complimentary cues across multiple burst frames to generate high-quality outputs. 
    Towards this goal, we develop a novel approach by solely focusing on the effective information exchange between burst frames, such that the degradations get filtered out while the actual scene details are preserved and enhanced. 
    Our central idea is to create a set of \emph{pseudo-burst} features that combine complimentary information from all the input burst frames to seamlessly exchange information. 
    %
    %
    However, the pseudo-burst cannot be successfully created unless the  individual burst frames are properly aligned to discount inter-frame movements. 
    Therefore, our approach initially extracts pre-processed features from each burst frame and matches them using an edge-boosting burst alignment module. 
    The pseudo-burst features are then created and enriched using multi-scale contextual information. 
    Our final step is to adaptively aggregate information from the pseudo-burst features to progressively increase resolution in multiple stages while merging the pseudo-burst features.
    In comparison to existing works that usually follow a late fusion scheme with single-stage upsampling, our approach performs favorably, delivering state-of-the-art performance on burst super-resolution, burst low-light image enhancement and burst denoising tasks. 
    The source code and pre-trained models are available at \url{https://github.com/akshaydudhane16/BIPNet}.
\end{abstract}


\begin{figure}[t]
    \centering
    \includegraphics[width=0.95\linewidth]{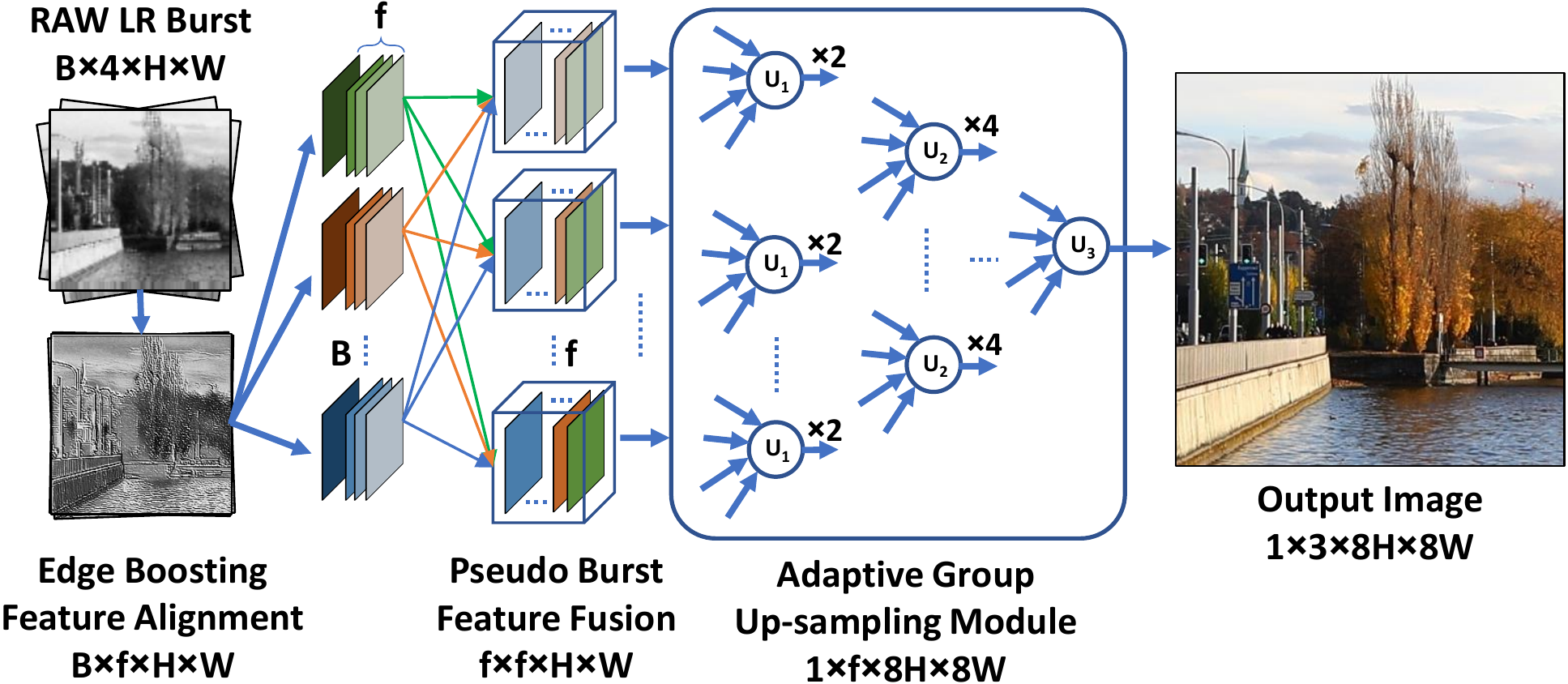}
    \vspace{-1mm}
    \caption{Holistic diagram of our burst image processing approach. Our network \xnet takes as input a RAW image burst and generates a high-quality RGB image. \xnet has three key stages. (1) Edge boosting feature alignment to remove noise, and inter-frame spatial and color misalignment. (2) Pseudo-burst feature fusion mechanism to enable inter-frame communication and feature consolidation. (3) Adaptive group upsampling to progressively increase spatial resolution while merging multi-frame information. While \xnet is generalizable to other restoration tasks, here we show super-resolution application.} 
    \label{fig: 1}
\end{figure}

\vspace{-2em}
\section{Introduction}
    High-end DSLR cameras can capture  images of excellent quality with vivid details. 
    With the growing popularity of smartphones, the main goal of computational photography is to generate DSLR-like images with smartphone cameras \cite{ignatov2017dslr}. 
    However, the physical constraints of smartphone cameras hinder the image reconstruction quality. 
    For instance, small sensor size limits spatial resolution and small lens and aperture provides noisy and color distorted images in low-light conditions \cite{delbracio2021mobile}. 
    Similarly, small pixel cavities accumulate less light therefore yielding low-dynamic range. 
    To alleviate these issues, burst (multi-frame) photography is a natural  solution instead of single-frame processing~\cite{hasinoff2016burst}.

    The goal of burst imaging is to composite a high-quality image by merging desired information from a collection of (degraded) frames of the same scene captured in a rapid succession. %
    However, burst image acquisition presents its own challenges. 
    For example, during image burst capturing, any movement in camera and/or scene objects will cause misalignment issues, thereby leading to ghosting and blurring artifacts in the output image~\cite{wronski2019handheld}. 
    Therefore, there is a pressing need to develop a multi-frame processing algorithm that is robust to alignment problems and requires no special burst acquisition conditions. 
    We note that existing burst processing techniques \cite{bhat2021deep, bhat2021deep1} extract and align features of burst images separately and usually employ late feature fusion mechanisms, which can hinder flexible information exchange among frames. 
    In this paper, we present a burst image processing approach, named \xnet, which is based on a novel pseudo-burst feature fusion mechanism that enables inter-frame communication and feature consolidation. 
    Specifically, a pseudo-burst is generated by exchanging information across frames such that each feature in the pseudo-burst contains complimentary properties of all input burst frames.

    Before synthesizing pseudo-bursts, it is essential to align the input burst frames (having arbitrary displacements) so that the relevant pixel-level cues are aggregated in the later stages. 
    Existing works~\cite{bhat2021deep,bhat2021deep1} generally use explicit motion estimation techniques (e.g., optical flow) to align input frames which are typically bulky pretrained modules that cannot be fully integrated within an end-to-end learnable pipeline. 
    This can result in errors caused during the flow estimation stage to be propagated to the warping and image processing stages, thereby negatively affecting the generated outputs. 
    In our case, the proposed \xnet implicitly learns the frame alignment with deformable convolutions~\cite{zhu2019deformable} that can effectively adapt to the given problem. 
    Further, we integrate the edge boosting refinement via back-projection operation~\cite{haris2018deep} in the alignment stage to retain high-frequency information. 
    It facilitates sustaining the alignment accuracy in cases where highly complex motions between burst images exist and only the deformable convolutional may not be sufficient for reliable alignment.

    Noise is always present in images irrespective of the lighting condition in which we acquire them. 
    Therefore one of our major goals is to remove noise~\cite{zamir2020cycleisp} early in the network to reduce difficulty for the alignment and fusion stages. 
    To this end, we incorporate residual global context attention in \xnet for feature extraction and refinement/denoising. 
    While the application of \xnet can be generalized to any burst processing task, we demonstrate its effectiveness on burst super-resolution, burst low-light image enhancement and burst denoising. 
    In super-resolution (SR), upsampling is the key step for image reconstruction. 
    Existing burst SR methods \cite{bhat2021deep, bhat2021deep1} first fuse the multi-frame features, and then use pixel-shuffle operation \cite{shi2016real} to obtain the high-resolution image.
    However, we can leverage the information available in multiple frames to perform merging and upsampling in a flexible and effective manner. 
    As such, we include adaptive group upsampling in our \xnet that progressively increases the resolution while merging complimentary features. \xnet schematic is shown in \cref{fig: 1}.

    \noindent The main contributions of this work include:\vspace{-0.5em}
        \begin{itemize}\setlength{\itemsep}{0em}
            \item An edge boosting alignment technique that removes spatial and color misalignment issues among the burst features. (Sec.~\ref{sec:EBFA})
            \item A novel pseudo-burst feature fusion mechanism to enable inter-frame communication and feature consolidation. (Sec.~\ref{sec:PBFF})
            \item An adaptive group upsampling module for progressive fusion and upscaling. (Sec.~\ref{sec:AGU})
        \end{itemize}
        Our \xnet achieves state-of-the-art results on synthetic and real benchmark datasets for the burst super-resolution, low-light image enhancement and burst denoising tasks. We provide comprehensive ablation results and visual examples to highlight the contributing factors in \xnet (Sec.~\ref{sec:experiments}).


 \section{Related Work}
 
     \noindent \textbf{Single Image Super-resolution (SISR).} 
        Since the first CNN-based work \cite{dong2014learning}, data-driven approaches have achieved high performance gains over the conventional counterparts \cite{yang2010image,freeman2002example}. 
         The success of CNNs is mainly attributed to their architecture design~\cite{anwar2020deep,zamir2021multi}. 
         Given a low-resolution image (LR), early methods learn to directly generate latent SR image~\cite{dong2014learning,dong2015image}. 
         In contrast, recent approaches learns to produce high frequency residual to which LR image is added to generate the final SR output~\cite{tai2017image,hui2018fast}. 
         Other notable SISR network designs employ recursive learning \cite{kim2016deeply,ahn2018fast}, progressive reconstruction \cite{wang2015deep,Lai2017}, attention mechanisms \cite{RCAN,dai2019second,zhang2020residual,zamir2020learning}, and generative adversarial networks \cite{wang2018esrgan, sajjadi2017enhancenet,SRResNet}. 
         The SISR approaches cannot handle multi-degraded frames from an input burst, while our approach belong to multi-frame SR that allows effectively merging cross-frame information towards a HR output. 
   
   \vspace{0.2em}
     \noindent \textbf{Multi-Frame Super-Resolution (MFSR).} Tsai \etal \cite{tsai1984multiframe} are the first to deal with the MFSR problem. 
         They propose a frequency domain based method that performs registration and fusion of the multiple aliased LR images to generate a SR image.
         Since processing multi-frames in the frequency domain leads to visual artifacts~\cite{tsai1984multiframe}, several other works aim to improve results by incorporating image priors in HR reconstruction process~\cite{stark1989high}, and making algorithmic choices such as iterative back-projection~\cite{peleg1987improving,irani1991improving}.
         Farsui \etal \cite{farsiu2004multiframe} design a joint multi-frame demosaicking and SR approach that is robust to noise. 
         MFSR methods are also developed for specific applications, such as for handheld devices~\cite{wronski2019handheld}, to increase spatial resolution of face images~\cite{ustinova2017deep}, and in satellite imagery~\cite{deudon2020highres,molini2019deepsum}.
         Lecouat \etal \cite{lecouat2021lucas} retains the interpretability of conventional approaches for inverse problems by introducing a deep-learning based optimization process that alternates between motion and HR image estimation steps. 
         Recently, Bhat \etal \cite{bhat2021deep} propose a multi-frame burst SR method that first aligns burst image features using an explicit PWCNet \cite{sun2018pwc} and then perform feature integration using an attention-based fusion mechanism. 
         However, explicit use of motion estimation and image warping techniques can pose difficulty handling scenes with fast object motions. 
         Recent works~\cite{tian2020tdan, wang2019edvr} show that the deformable convolution~\cite{zhu2019deformable} effectively handles inter-frame alignment issues due to being implicit and adaptive in nature. 
         Unlike existing MFSR methods, we implicitly learn the inter-frame alignment and then channel-wise aggregate information followed by adaptive upsampling to effectively utilize multi-frame information. 

\begin{figure*}[t]
    \centering
    \includegraphics[width=0.95\linewidth]{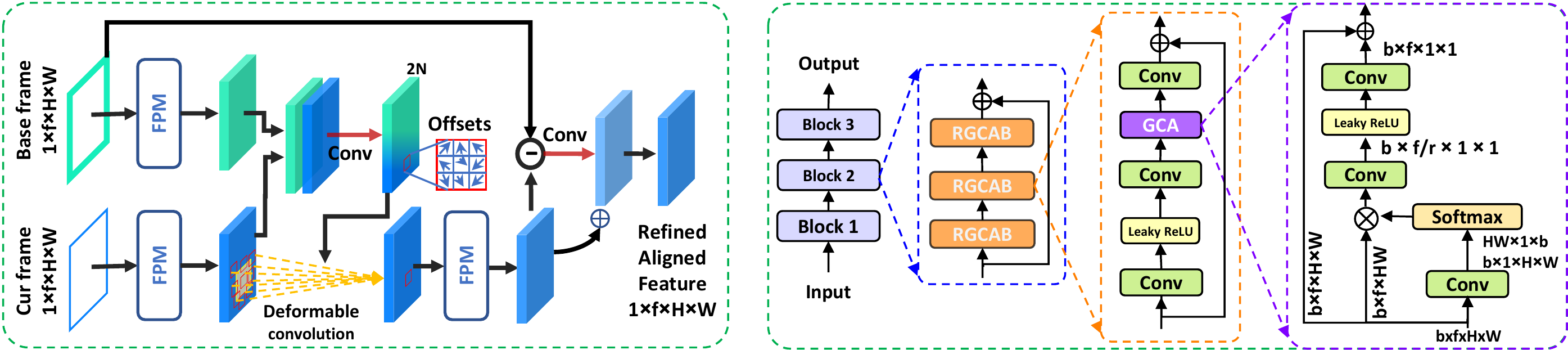}
    
    \caption{Edge boosting feature alignment (EBFA) module aligns all other images in the input burst to the base frame.  Feature processing module (FPM) is added in EBFA to denoise input frames to facilitate the easy alignment. $\otimes$ represents matrix multiplication.} 
    \label{fig: 2}
\end{figure*}

    \vspace{0.2em}
  \noindent   \textbf{Low-Light Image Enhancement.} 
         Images captured in low-light conditions are usually dark, noisy and color distorted. 
         These problems are somewhat alleviated by using long sensor exposure time, wide aperture, camera flash, and exposure bracketing \cite{delbracio2021mobile, zamir2021learning}. 
         However, each of these solutions come with their own challenges. 
         For example, long exposure yields images with ghosting artifacts due to camera or object movements. 
         Wide apertures are not available on smartphone devices, etc. 
         See-in-the-Dark method~\cite{chen2018learning} is the first attempt to replace the standard camera imaging pipeline with a CNN model. 
         It takes as input a RAW input image captured in extreme low-light and learns to generate a well-lit sRGB image. 
         Later this work is further improved with a new CNN-based architecture~\cite{maharjan2019improving} and by employing a combined pixel-wise and perceptual loss \cite{zamir2021learning}. 
         Zaho \etal \cite{zhao2019end} takes the advantage of burst imaging and propose a recurrent convolutional network that can produce noise-free bright sRGB image from a burst of RAW images.
         The results are further improved by Karadeniz \etal \cite{karadeniz2020burst} with their two-stage approach: first sub-network performs denoising, and the second sub-network provides visually enhanced image. 
         Although these studies demonstrate significant progress in enhancing low-light images, they do not address inter-frame misalignment and information interaction which we address in this work.

\vspace{0.2em}
   \noindent \textbf{Multi-Frame Denoising.}
        Earlier works \cite{Dabov2007VideoDB,Maggioni2011VideoDU,Maggioni2012VideoDD} extend the popular image denoising algorithm BM3D~\cite{Dabov2007ImageDB} to video. Buades \etal perform denoising by estimating the noise level from the aligned images followed by the combination of pixel-wise mean and BM3D. A hybrid 2D/3D Wiener filter is used in~\cite{Hasinoff2016BurstPF} to denoise and merge burst images for high dynamic range and low-light photography tasks. Godard~\etal~\cite{Godard2018DeepBD} utilize recurrent neural network (RNN) and extend a single image denoising network for multiple frames. Mildenhall~\etal~\cite{Mildenhall2018BurstDW} generate per-pixel kernels through the kernel prediction network (KPN) to merge the input images. In~\cite{Marinc2019MultiKernelPN}, authors extend KPN approach to predict multiple kernels, while~\cite{Xia2020BasisPN} introduce basis prediction networks (BPN) to enable the use of larger kernels. Recently, Bhat~\etal~\cite{bhat2021deep1} propose a deep reparameterization of the maximum a posteriori formulation for the multi-frame SR and denoising.

            
\section{Burst Processing Approach}
    In this section, we describe our burst processing approach which is applicable to different image restoration tasks, including burst super-resolution, burst low-light image enhancement and burst denoising. 
    The goal is to generate a high-quality image by combining information from multiple degraded images captured in a single burst.  
    Burst images are typically captured with handheld devices, and it is often inevitable to avoid inter-frame spatial and color misalignment issues. 
    Therefore, the main challenge of burst processing is to accurately align the burst frames, followed by combining their complimentary information while preserving and reinforcing the shared attributes. 
    To this end, we propose  \xnet in which different modules operate in synergy to jointly perform denoising, demosaicking, feature fusion, and upsampling tasks in a unified model.

\vspace{0.2em}    
  \noindent  \textbf{Overall pipeline.} Fig.~\ref{fig: 1} shows three main stages in the proposed \xnet. 
    First, the input RAW burst is passed through the edge boosting feature alignment module to extract features, reduce noise, and remove spatial and color misalignment issues among the burst features (Sec.~\ref{sec:EBFA}). 
    Second, a pseudo-burst is generated by exchanging information such that each feature map in the pseudo-burst now contains complimentary properties of all actual burst image features (Sec.~\ref{sec:PBFF}). 
    Finally, the multi-frame pseudo-burst features are processed with the adaptive group upsampling module to produce the final high-quality image (Sec.~\ref{sec:AGU}). 

    \subsection{Edge Boosting Feature Alignment Module}\label{sec:EBFA}
        One major challenge in burst processing is to extract features from multiple degraded images that are often contaminated with noise, unknown spatial displacements, and color shifts. 
        These issues arise due to camera and/or object motion in the scene, and lighting conditions. 
        To align the other images in the burst with the base frame (usually the $1^{st}$ frame for simplicity) we propose an alignment module based on modulated deformable convolutions \cite{zhu2019deformable}. 
        However, existing deformable convolution is not explicitly designed to handle noisy RAW data.
        Therefore, we propose a feature processing module to reduce noise in the initial burst features.
        Our edge boosting feature alignment (EBFA) module (Fig.~\ref{fig: 2}(a)) consists of  feature processing followed by burst feature alignment.
        
        \subsubsection{Feature Processing Module}\label{sec:FPM}
            The proposed feature processing module (FPM), shown in Fig.~\ref{fig: 2}(b), employs residual-in-residual learning that allows abundant low-frequency information to pass easily via skip connections~\cite{RCAN}.
            Since capturing long-range pixel dependencies which extracts global scene properties has been shown to be beneficial for a wide range of image restoration tasks~\cite{zamir2021restormer} (e.g., image/video super-resolution \cite{mei2020image} and extreme low-light image enhancement \cite{arora2021low}), we utilize a global context attention (GCA) mechanism to refine the latent representation produced by residual block, as illustrated in Fig.~\ref{fig: 2}(b). 
            Let $\left\{ {{\bm{x}^b}} \right\}_{{b \in [1:B]}} {\in} \mathbb{R}^{B \times f \times H \times W}$ be an initial latent representation of the burst having $B$ burst images and $f$ number of feature channels,  our residual global context attention block (RGCAB in Fig.~\ref{fig: 2}(b)) is defined as:
            \begin{equation}
                {\bm{y}^b} = {\bm{x}^b} + {W_{1}}\left( {\alpha \left( {{{\bm{\bar x}}^b}} \right)} \right),
            \end{equation}
            where 
            ${{\bm{\bar x}}^b} = {W_{3}}\left( {\gamma \left( {{W_{3}}\left( {{\bm{x}^b}} \right)} \right)} \right)$ and 
            $\alpha \left( {{{\bm{\bar x}}^b}} \right) = {\bm{\bar x^b}} + {W_{1}}\left( {\gamma \left( {{W_{1}}\left( {\Psi \left( {{W_{1}}\left( {{{\bm{\bar x}}^b}} \right)} \right) \otimes {{\bm{\bar x}}^b}} \right)} \right)} \right)$.
            Here, $W_{k}$ represents a convolutional layer with $k \times k$ sized filters and each $W_{k}$ corresponds to a separate layer with distinct parameters, $\gamma$ denotes leaky ReLU activation, $\Psi$ is softmax activation, $\otimes$ represents matrix multiplication, and $\alpha(\cdot)$ is the global context attention. 
        
        \begin{figure*}[t]
          \centering
          \includegraphics[width=0.8\textwidth]{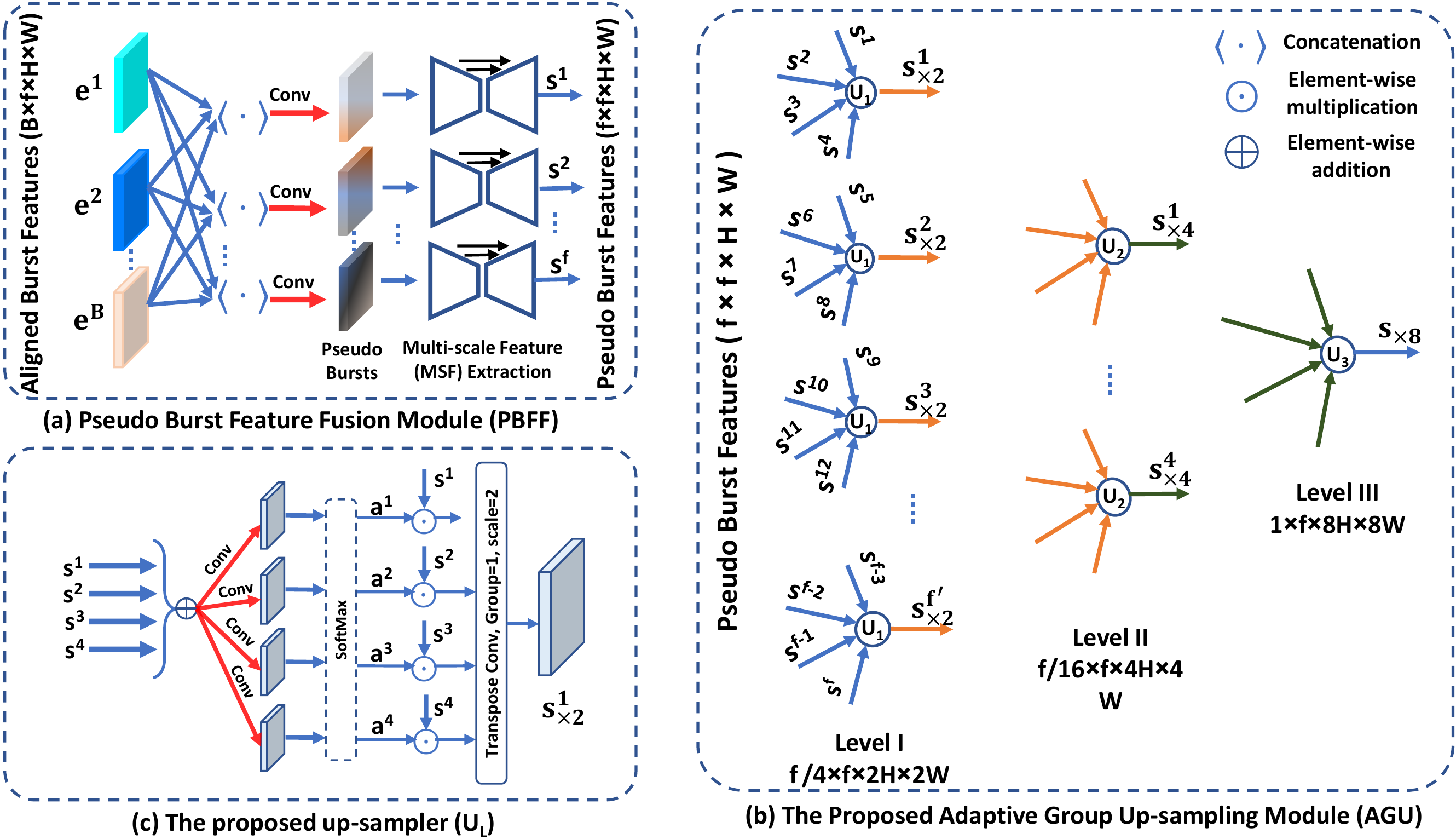}
          \vspace{-2mm}
          \caption{(a) Pseudo-burst is generated by exchanging information across frames such that each feature tensor in the pseudo-burst contains complimentary properties of all frames. Pseudo bursts are processed with (shared) U-Net to extract multi-scale features. (b) AGU module handles pseudo-bursts features in groups and progressively performs upscaling. (c) Schematic of dense-attention based upsampler.} 
          \label{fig: 3}
        \end{figure*}

        \subsubsection{Burst Feature Alignment Module}\label{sec:DAM}
            To effectively fuse information from multiple frames, these frame-level features need to be aligned first. 
            We align the features of the current frame $\bm{y}^{b}$ with the base frame\footnote{In this work, we consider first input burst image as the base frame.} $\bm{y}^{b_r}$. EBFA processes $\bm{y}^b$ and $\bm{y}^{b_r}$ through an offset convolution layer and predicts the offset $\Delta {n}$ and modulation scalar $\Delta m$ values for $\bm{y}^b$. 
            The aligned features ${\bm{\bar y}}^b$ computed as:
            \begin{equation}
                \quad {{\bm{\bar y}}^b} = {W^d}\left( {{\bm{y}^b},\;\Delta {n},\;\Delta m} \right),\;
                \Delta m = {W^{o}}\left( {{\bm{y}^b},\;{\bm{y}^{{b_r}}}} \right),
            \end{equation}
            where, $W^d$ and $W^{o}$ represent the deformable and offset convolutions, respectively. More specifically, each position $n$ on the aligned feature map ${\bm{\bar y}}^b$ is obtained as:
            \begin{equation}
                {{\bm{\bar y}}^b}_n = \sum\limits_{i=1}^K {{W_{n_i}^{d}}\,\,\,{\bm{y}_{\left( n + {n_i} + \Delta {n_i} \right)}^b}} \cdot \Delta {m_{n_i}},
                 \vspace{-0.8em}
            \end{equation}
            where, $K$=9, $\Delta$m lies in the range [0, 1] for each $n_i \in \left\{ { (-1, 1),  \allowbreak (-1, 0), ..., (1,1)} \right\}$ is a regular grid of $3{\times}3$ kernel.
            
            The convolution operation will be performed on the non-uniform positions $({n_i} + \Delta {n_i})$, where ${n_i}$ can be fractional. To avoid fractional values, the operation is implemented using bilinear interpolation.
            
            The proposed EBFA module is inspired from the deformable alignment module (DAM) \cite{tian2020tdan} with the following differences. Our approach does not provide explicit ground-truth supervision to the alignment module, instead it learns to perform implicit alignment. Furthermore, to strengthen the feature alignment and to correct the minor alignment errors, using FPM, we obtain refined aligned features (RAF) followed by computing the high-frequency residue by taking the difference between the RAF and base frame features and add it to the RAF. The overall process of our EBFA module is summarized as: ${\bm{e}}^b = {\bm{\bar y}}^b + W_{3}\left( {{\bm{\bar y}}^b} - {\bm{y}^{{b_r}}} \right)$
            where ${{\bm{e}}^b} \in \mathbb{R}^{B \times f \times H \times W}$ represents the aligned burst feature maps, and $W_{3}(\cdot)$ is the convolution. Although the deformable convolution is shown only once in Fig.~\ref{fig: 2}(a) for brevity, we sequentially apply three such layers to improve the transformation capability of our EBFA module.

    \subsection{Pseudo-Burst Feature Fusion Module}\label{sec:PBFF}
        Existing burst image processing techniques \cite{bhat2021deep, bhat2021deep1} separately extract and align features of burst images and usually employ late feature fusion mechanisms, which can hinder flexible information exchange between frames. 
        We instead propose a pseudo-burst feature fusion (PBFF) mechanism (see Fig.~\ref{fig: 3} (a)). 
        This PBFF module generates feature tensors by concatenating the corresponding channel-wise features from all burst feature maps. 
        Consequently, each feature tensor in the pseudo-burst contains complimentary properties of all actual burst image features. 
        Processing inter-burst feature responses simplifies the representation learning task and merges the relevant information by decoupling the burst image feature channels. 
        Given the aligned burst feature set  ${e} = \left\{ {\bm{e}_{c}^b} \right\}_{c \in [1:{f}]}^{b \in [1:B]}$ of burst size $B$ and $f$ number of channels, the pseudo-burst is generated by,
        \begin{equation}
            \bm{S}^c = {W^\rho}\left( {\left\langle {\bm{e}_{c}^{{1}},\;\bm{e}_{c}^{{2}},\; \cdots ,\;\bm{e}_{c}^{{B}}} \right\rangle } \right), \quad s. t. \quad c \in [1:f],
        \end{equation}
        where, $\left\langle \cdot \right\rangle$ represents concatenation, ${\bm{e}_{c}^{{1}}}$ is the ${c}^{th}$ feature map of $1^{st}$ aligned burst feature set ${\bm{e}^1}$, $W^\rho$ is the convolution layer with $f$ output channel, and ${\bm{S}} = \left\{ {\bm{S}^c} \right\}_{c \in [1:{f}]}$ represents the pseudo-burst of size $f\times f\times H\times W$. 
        In this paper, we use $f=64$.
        
        Even after generating pseudo-bursts, obtaining their deep representation is essential. We use a light-weight (3-level) U-Net to extract multi-scale features (MSF) from pseudo-bursts. We use shared weights in the U-Net, and also employ our FPM instead of regular convolutions.
        
    \subsection{Adaptive Group Upsampling Module}\label{sec:AGU}
        Upsampling is the final key step to generate the super-resolved image from LR feature maps. Existing burst SR methods \cite{bhat2021deep, bhat2021deep1} use pixel-shuffle layer \cite{shi2016real} to perform upsampling in one-stage. However, in burst image processing, information available in multiple frames can be  exploited effectively to get into HR space. To this end, we propose to \emph{adaptively} and \emph{progressively} merge multiple LR features in the upsampling stage. 
        For instance, on the one hand it is beneficial to have uniform fusion weights for texture-less regions in order to perform denoising among the frames. On the other hand, to prevent ghosting artifacts, it is desirable to have low fusion weights for any misaligned frame. 
        
        Fig.~\ref{fig: 3}(b) shows the proposed adaptive group upsampling (AGU) module that takes as input the feature maps ${\bm{S}} = \left\{ {\bm{S}^c} \right\}_{c \in [1:{f}]}$ produced by the pseudo-burst fusion module and provides a super-resolved output via three-level progressive upsampling. In AGU, we  sequentially divide the pseudo-burst features into groups of 4, instead of following any complex selection mechanism.  
        These groups of features are upsampled with the architecture depicted in Fig.~\ref{fig: 3}(c) that first computes a dense attention map ($\bm{a}^c$), carrying attention weights for each pixel location. The dense attention maps are element-wise applied to the respective burst features.
        Finally, the upsampled response for a given group of features ${\hat{\bm{S}}^{g}} = \left\{ {\bm{S}^{i} : i \in [ (g-1)*4+1 : g*4] } \right\}^{g \in [1:f/4]} \subset \bm{S}$ and associated attention maps $\hat{\bm{a}}^g$ at the first upsampling level (Level I in Fig.~\ref{fig: 3}(b)) is formulated as:
        \begin{align}
            \bm{S}^g_{\times 2} &= {W_T}\left( {\left\langle { {{\hat{\bm{S}}^{{g}}} \odot {\hat{\bm{a}}^{{g}}}} } \right\rangle } \right),  \notag  \\
            {\hat{\bm{a}}^{{g}}} &= \psi \Bigg( {W_1 \Bigg( {{W_1} \Bigg( {\sum\limits_{i = (g-1)*4+1}^{g*4} {{\bm{S}^{i}}} } \Bigg)} \Bigg)} \Bigg),
        \end{align}
        where $\psi\left(\cdot\right)$ denotes the softmax activation function, $W_T$ is the $3 \times 3$  Transposed convolution layer, and $\hat{\bm{a}}^{{g}} \in \mathbb{R}^{4\times f \times H \times W}$ represents the dense attention map for $g^{th}$ burst feature response group (${\hat{\bm{S}}^g}$).

        To perform burst SR of scale factor $\times 4$, we need in fact $\times 8$ upsampling (additional $\times 2$ is due to the mosaicked RAW LR frames). Thus, in AGU we employ three levels of $\times2$ upsampling. As, our BIPNet generates 64 pseudo bursts, this naturally forms 16, 4 and 1 feature groups at levels I, II, and III, respectively. Upsampler at each level is shared among groups to avoid the increase in network parameters.

\section{Experiments}\label{sec:experiments}
    We evaluate the proposed \xnet and other state-of-the-art approaches on real and synthetic datasets for  \textbf{(a)} burst super-resolution, \textbf{(b)} burst low-light image enhancement, and \textbf{(c)} burst denoising.

 \noindent   \textbf{Implementation Details.}
        Our \xnet is end-to-end trainable and needs no pretraining of any module. For network parameter efficiency, all burst frames are processed with shared \xnet modules (FPM, EBFA, PBFF and AGU).
        Overall, the proposed network contains 6.67M parameters.
        We train separate model for burst SR, burst low-light image enhancement and burst denoising using $L_1$ loss only. While for SR on real data, we fine-tune our BIPNet with pre-trained weights on SyntheticBurst dataset using aligned $L_1$ loss~\cite{bhat2021deep}. The models are trained with Adam optimizer. 
    Cosine annealing strategy \cite{loshchilov2016sgdr} is employed to steadily decrease the learning rate from $10^{-4}$ to $10^{-6}$ during training. We use horizontal and vertical flips for data augmentation. Additional network details and visual results are provided in the supplementary material.

    \subsection{Burst Super-resolution}
        We perform SR experiments for scale factor $\times$4 on the SyntheticBurst and (real-world) BurstSR datasets~\cite{bhat2021deep}

        \noindent \textbf{Datasets.} \textbf{(1) SyntheticBurst} dataset consists of 46,839 RAW bursts for training and 300 for validation. 
        Each burst contains 14 LR RAW images (each of size $48{\times}48$ pixels) that are synthetically generated from a single sRGB image. 
        Each sRGB image is first converted to the RAW space using the inverse camera pipeline~\cite{brooks2019unprocessing}. 
        Next, the burst is generated with random rotations and translations. %
        Finally, the LR burst is obtained by applying the bilinear downsampling followed by Bayer mosaicking, sampling and random noise addition operations. 
        \textbf{(2) BurstSR} dataset consists of 200 RAW bursts, each containing 14 images. To gather these burst sequences, the LR images and the corresponding (ground-truth) HR images are captured with a smartphone camera and a DSLR camera, respectively. 
        From  200 bursts, 5,405 patches are cropped for training and 882 for validation. Each input crop is of size $80{\times}80$ pixels.

        \vspace{0.2em}
        \noindent \textbf{SR results on synthetic data.}
            The proposed \xnet is trained for 300 epochs on training set while evaluated on validation set of SyntheticBurst dataset~\cite{bhat2021deep}. We compare our \xnet with the several burst SR method such as HighResNet~\cite{deudon2020highres}, DBSR~\cite{bhat2021deep}, LKR~\cite{lecouat2021lucas}, and MFIR~\cite{bhat2021deep1} for $\times$4 upsampling. Table~\ref{tab: sr} shows that our method performs favorably well. Specifically, our \xnet achieves PSNR gain of 0.37 dB over the previous best method MFIR~\cite{bhat2021deep1} and 0.48 dB over the second best approach \cite{lecouat2021lucas}.
            
\begin{figure}[t]
        \centering
        \includegraphics[width=1\linewidth]{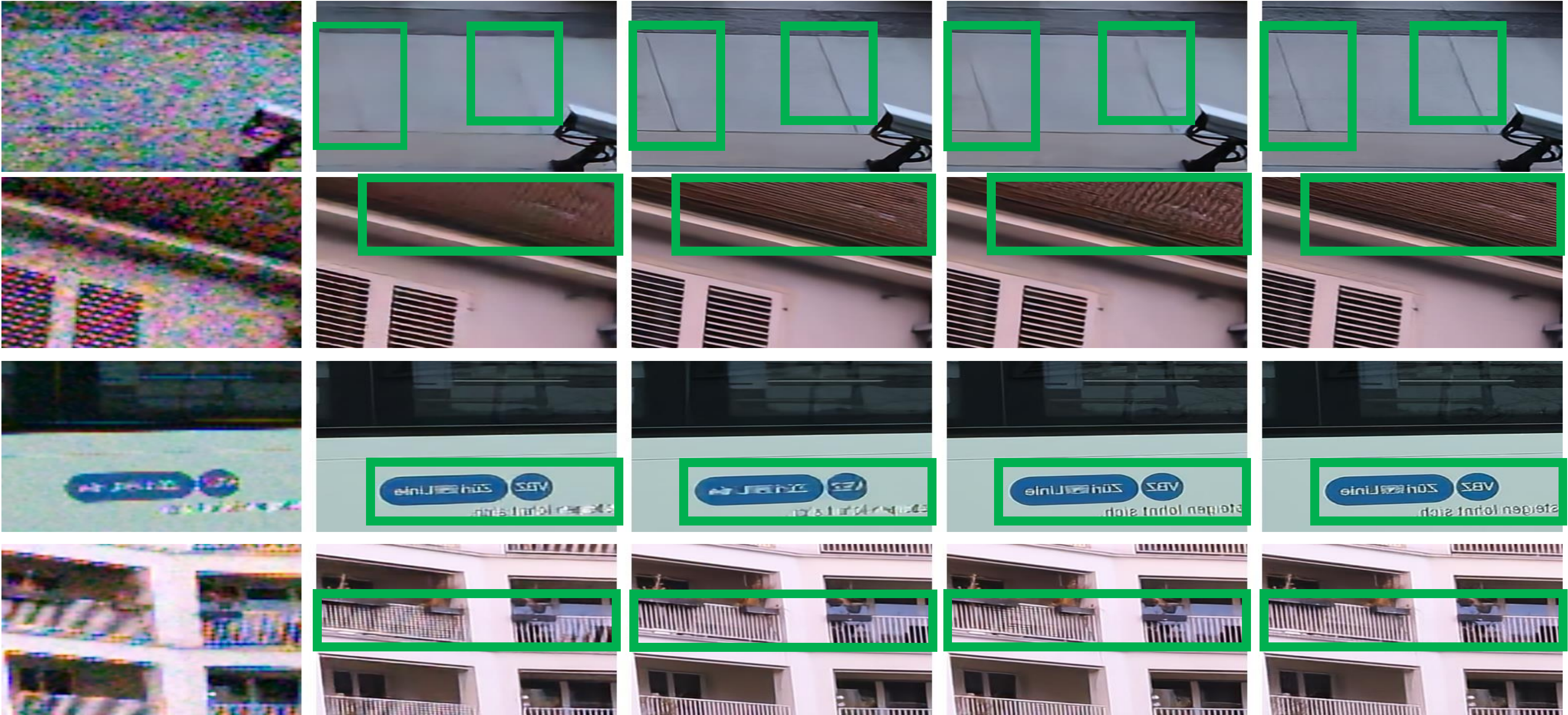}
        \scriptsize \,\,\, Base frame \quad \qquad DBSR~\cite{bhat2021deep} \qquad LKR~\cite{lecouat2021lucas} \quad \qquad MFIR~\cite{bhat2021deep1}  \qquad \xnet (ours)
        \vspace{-2mm}
        \caption{Comparisons for $\times4$ burst SR on SyntheticBurst~\cite{bhat2021deep}. Our \xnet produces more sharper and clean results than other competing approaches.}
        \label{fig: 4}
    \end{figure}

            Visual results provided in Fig.~\ref{fig: 4} show that the SR images produced by \xnet are more sharper and faithful than those of the other algorithms. 
            Our \xnet is capable of reconstructing structural content and fine textures, without introducing artifacts and color distortions. Whereas, the results of DBSR, LKR and MFIR contain splotchy textures and compromise image details.
            
\begin{figure}[t]
        \centering
        \includegraphics[width=1\linewidth]{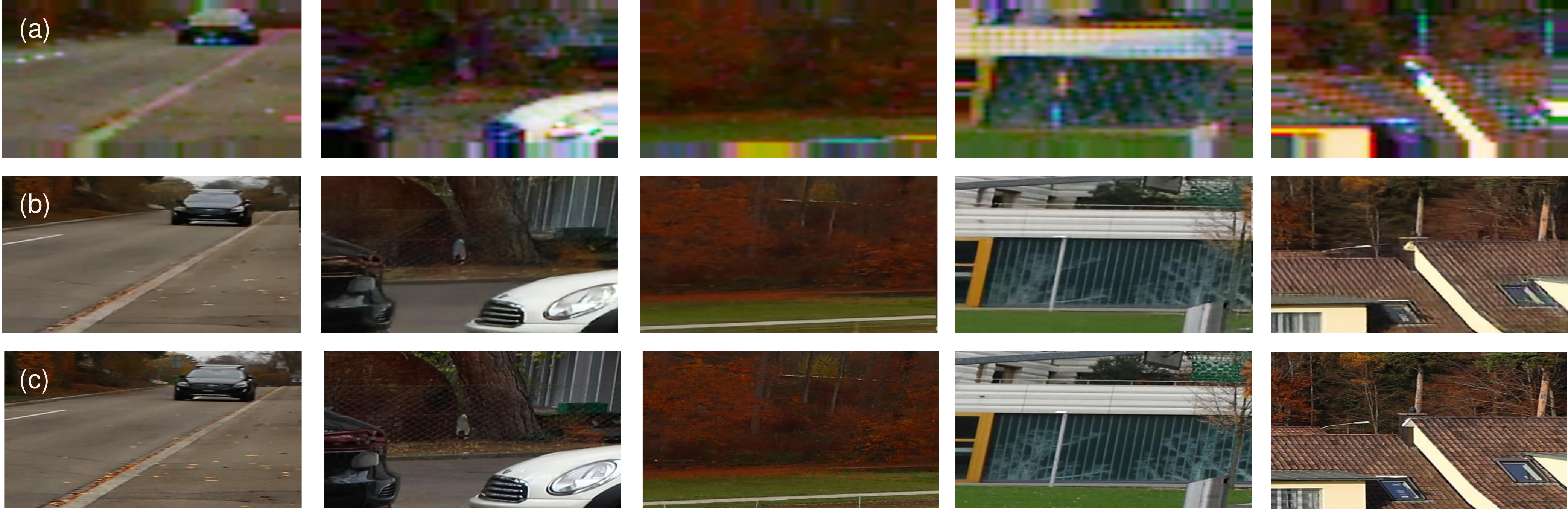}
        \vspace{-6mm}
        \caption{Results for  $\times$8 burst SR on SyntheticBurst dataset~\cite{bhat2021deep}. (a) Base frame (b) BIPNet (Ours) (c) Ground truth. Our method effectively recovers image details in extremely challenging cases.}
        \label{fig: x8 sr results}
    \end{figure}

            To show the effectiveness of our method \xnet on large scale factor, we perform experiments for the $\times8$ burst SR. We synthetically generate LR-HR pairs following the same procedure as we described above for the SyntheticBurst dataset. Visual results in Fig. \ref{fig: x8 sr results} show that our \xnet is capable of recovering rich details for such large scale factors as well, without any artifacts. Additional examples can be found in supplementary material.

     \begin{table}[t]
            \centering
            \caption{Performance evaluation on synthetic and real burst validation sets \cite{bhat2021deep} for $\times$4 burst super-resolution.}
            \vspace{-1mm}
            \label{tab: sr}
            \setlength{\tabcolsep}{2pt}
            \scalebox{0.85}{
                \begin{tabular}{l@{$\;\,$}c@{$\;\,$}c@{$\;\,$}c@{$\;\,$}c}
                    \toprule
                        \multirow{2}{4em}{\textbf{Methods}}  & \multicolumn{2}{c}{\textbf{SyntheticBurst}} & \multicolumn{2}{c}{\textbf{(Real) BurstSR}} \\
                          \cmidrule{2-5}
                          & \textbf{PSNR $\uparrow$} & \textbf{SSIM $\uparrow$} & \textbf{PSNR $\uparrow$} & \textbf{SSIM $\uparrow$} \\
                    \midrule
                    Single Image &    36.17   &   0.909   &  46.29     &  0.982 \\
                    HighRes-net \cite{deudon2020highres} & 37.45 & 0.92 & 46.64 & 0.980 \\
                    DBSR~\cite{bhat2021deep} & 40.76 & 0.96 & 48.05 & 0.984 \\
                    LKR~\cite{lecouat2021lucas} & 41.45 & 0.95  & - & - \\
                    MFIR~\cite{bhat2021deep1} & 41.56 & 0.96 & 48.33 & 0.985 \\
                    \midrule
                    \textbf{\xnet~(Ours)} & \textbf{41.93} & \textbf{0.96} & \textbf{48.49} & \textbf{0.985} \\
                    \bottomrule
                \end{tabular}}
    \end{table}

        \begin{table*}[t]
            \parbox{.38\linewidth}{
                 \centering
                 \footnotesize
                 \caption{Importance of \xnet modules evaluated on SyntheticBurst validation set for $\times$4 burst SR.}
                  \vspace{-2mm}
                 \label{tab: ablations}
                 \setlength{\tabcolsep}{3pt}
                  \scalebox{0.75}{
                 \begin{tabular}{lcccccccc}
                     \toprule
                     \textbf{Modules} & \textbf{A1} & \textbf{A2} & \textbf{A3} & \textbf{A4} & \textbf{A5} & \textbf{A6} & \textbf{A7} & \textbf{A8} \\
                     \toprule
                     \toprule
                     {{Baseline}} & \ding{51}     & \ding{51}     & \ding{51}     & \ding{51}     & \ding{51}     & \ding{51}     & \ding{51}     & \ding{51} \\
                     {{FPM (\textsection\ref{sec:FPM})}} &       & \ding{51}     & \ding{51}     & \ding{51}     & \ding{51}     & \ding{51}     & \ding{51}     & \ding{51} \\
                     {{DAM (\textsection\ref{sec:DAM})}} &       &       & \ding{51}     & \ding{51}     & \ding{51}     & \ding{51}     & \ding{51}     & \ding{51} \\
                     {RAF (\textsection\ref{sec:DAM})} &       &       &       & \ding{51}     & \ding{51}     & \ding{51}     & \ding{51}     & \ding{51} \\
                     {PBFF (\textsection\ref{sec:PBFF})} &       &       &       &       & \ding{51}     & \ding{51}     & \ding{51}     & \ding{51} \\
                     {MSF (\textsection\ref{sec:PBFF})} &       &       &       &       &       & \ding{51}     & \ding{51}     & \ding{51} \\
                     {AGU (\textsection\ref{sec:AGU})} &       &       &       &       &       &       & \ding{51}     & \ding{51} \\
                     {EBFA (\textsection\ref{sec:EBFA})} &       &       &       &       &       &       &       & \ding{51} \\
                     \midrule
                     \textbf{PSNR} & {36.38} & {36.54} & {38.39} &  {39.10} &  {39.64} & {40.35} &  {41.25} & {41.55} \\
                     \bottomrule
                 \end{tabular}%
                 }
                }
            \hfill
            \parbox{.29\linewidth}{
                \centering
                \caption{Importance of the proposed alignment and fusion module evaluated on SyntheticBurstSR dataset for $\times4$ SR.}
                \vspace{-2mm}
                    \label{tab: ablation2}
                \setlength{\tabcolsep}{6pt}
                \scalebox{0.65}{
                    \begin{tabular}{ll@{$\;\,$}c@{$\;\,$}c@{$\;\,$}}
                        \toprule
                            & \textbf{Methods}  & \textbf{PSNR $\uparrow$} & \textbf{SSIM $\uparrow$}\\
                        \midrule
                        \multirow{3}[0]{*}{\textbf{(a)} Alignment} & Explicit~\cite{bhat2021deep} & 39.26 & 0.944\\
                        & TDAN~\cite{tian2020tdan} & 40.19 & 0.957 \\
                        & EDVR~\cite{wang2019edvr} & 40.46 & 0.958 \\
                        \midrule
                        \multirow{3}[0]{*}{\textbf{(b)} Fusion } & Addition & 39.18 & 0.943 \\
                        & Concat & 40.13 & 0.956 \\
                        & DBSR~\cite{bhat2021deep} & 40.16 & 0.957 \\
                        \midrule
                        \textbf{(c)} & \textbf{\xnet~(Ours)} & \textbf{41.55} & \textbf{0.96} \\
                        \bottomrule
                    \end{tabular}
                    }
                } \hfill
                \parbox{.29\linewidth}{
                \centering
            \caption{Burst low-light image enhancement methods evaluated on the SID dataset \cite{chen2018learning}. Our \xnet advances state-of-the-art by 3.07 dB.}
             \label{tab: enhancement}
             \vspace{-2mm}
            \setlength{\tabcolsep}{8pt}
            \scalebox{0.65}{
                \begin{tabular}{l@{$\;\,$}c@{$\;\,$}c@{$\;\,$}c@{$\;\,$}}
                    \toprule
                    \textbf{Methods} & \textbf{PSNR $\uparrow$} & \textbf{SSIM $\uparrow$} & \textbf{LPIPS $\downarrow$} \\
                    \midrule
                        Chen \etal \cite{chen2018learning} & 29.38 & 0.892 & 0.484 \\
                        Maharjan \etal \cite{maharjan2019improving} & 29.57 & 0.891 & 0.484 \\
                        Zamir \etal \cite{zamir2021learning} & 29.13 & 0.881 & 0.462 \\
                        Zhao \etal \cite{zhao2019end} & 29.49 & 0.895 & 0.455 \\
                        Karadeniz \etal \cite{karadeniz2020burst} & 29.80 & 0.891 & 0.306 \\
                        \midrule
                        \textbf{\xnet~(Ours)} & \textbf{32.87} & \textbf{0.936} & \textbf{0.305} \\
                    \bottomrule
                \end{tabular}}
                }
        \end{table*}

\begin{figure}[t]
        \centering
        \includegraphics[width=1\linewidth]{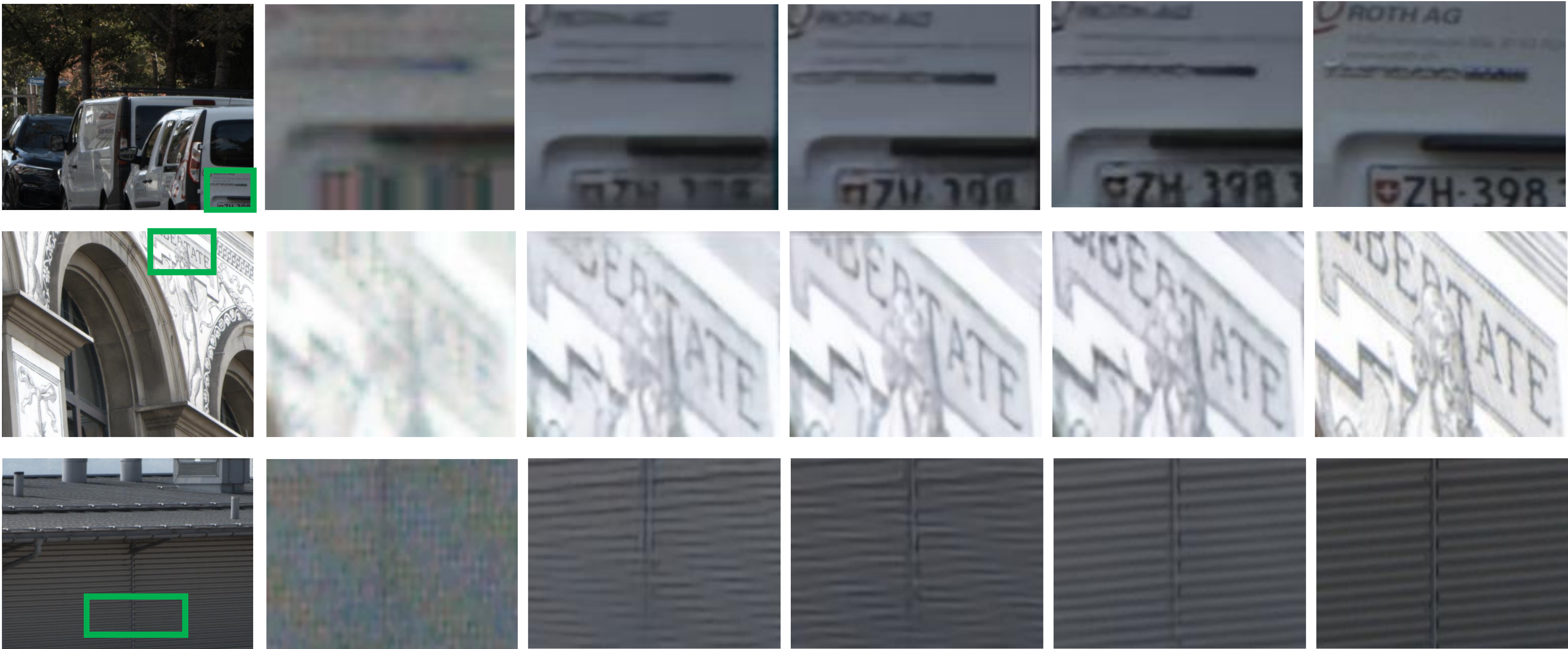}
        \scriptsize HR Image \quad Base frame \quad DBSR~\cite{bhat2021deep} \quad MFIR~\cite{bhat2021deep1}  \quad \xnet (ours) \qquad GT
        \vspace{-2mm}
        \caption{Comparisons for $\times4$ burst super-resolution on Real BurstSR dataset~\cite{bhat2021deep}. Our \xnet produces more sharper and clean results than other competing approaches.}
        \label{fig: sr results}
    \end{figure}

        \vspace{0.2em}
        \noindent \textbf{SR results on real data.}
            The LR input bursts and the corresponding HR ground-truth in BurstSR dataset suffer with minor misalignment as they are captured with different cameras. To mitigate this issue, we use aligned L1 loss for training and aligned PSNR/SSIM for evaluating our model, as in previous works~\cite{bhat2021deep,bhat2021deep1}. We fine-tuned the pre-trained \xnet for 15 epochs on training set while evaluated on validation set of BurstSR dataset. 
            The image quality scores are reported in Table \ref{tab: sr}. Compared to the previous best approach MFIR~\cite{bhat2021deep1}, our \xnet provides performance gain of 0.16 dB. The visual comparisons in Fig.~\ref{fig: sr results} show that our \xnet is more effective in recovering fine details in the reproduced images than other competing approaches.
    
        \vspace{0.2em}
            \noindent \textbf{Ablation Study.}
             Here we present ablation experiments to demonstrate the impact of each individual component of our approach. All ablation models are trained for 100 epochs on SyntheticBurst dataset~\cite{bhat2021deep1} for SR scale factor $\times4$. Results are reported in Table~\ref{tab: ablations}. 
             For the baseline model, we employ Resblocks~\cite{lim2017edsr} for feature extraction, simple concatenation operation for fusion, and transpose convolution for upsampling. 
             The baseline network achieves 36.38 dB PSNR. When we add the proposed modules to the baseline, the results improve significantly and consistently. 
             For example,  we obtain performance boost of 1.85 dB when we consider the deformable alignment module DAM. 
             Similarly, RAF contributes 0.71 dB improvement towards the model. With our PBFF mechanism, the network achieves significant gain of 1.25 dB. 
             AGU brings 1 dB increment in the upsampling stage. 
             Finally, EBFA demonstrates its effectiveness in correcting alignment errors by providing 0.3 dB improvement in PSNR. 
             Overall, our \xnet obtains a compelling gain of 5.17 dB over the baseline method.

             Finally, we perform ablation experiments to demonstrate the importance of the proposed EBFA and PBFF modules by replacing them with existing alignment and fusion modules. Table~\ref{tab: ablation2}(a) shows that replacing our EBFA with other alignment modules have negative impact (PSNR drops at least over $1$ dB).  Similar trend can be observed when using fusion strategies other than our PBFF, see Table~\ref{tab: ablation2}(b).

    \subsection{Burst Low-Light Image Enhancement}
        To further demonstrate the effectiveness of \xnet, we perform experiments for burst low-light image enhancement. Given a low-light RAW burst, our goal is to generate a well-lit sRGB image. Since the input is mosaicked RAW burst, we use one level AGU to obtain the output.
    
    \vspace{0.2em}
    \noindent \textbf{Dataset.} {SID} dataset~\cite{chen2018learning} consists of input RAW burst images captured with short-camera exposure in low-light conditions, and their corresponding ground-truth sRGB images. The Sony subset contains 161, 20 and 50 distinct burst sequences for training, validation and testing, respectively. We prepare 28k patches of spatial size $128 \times 128$ with burst size 8 from the training set of Sony subset of SID to train the network for 50 epochs. 
        
\begin{figure}[t]
        \centering
        \includegraphics[width=1\linewidth]{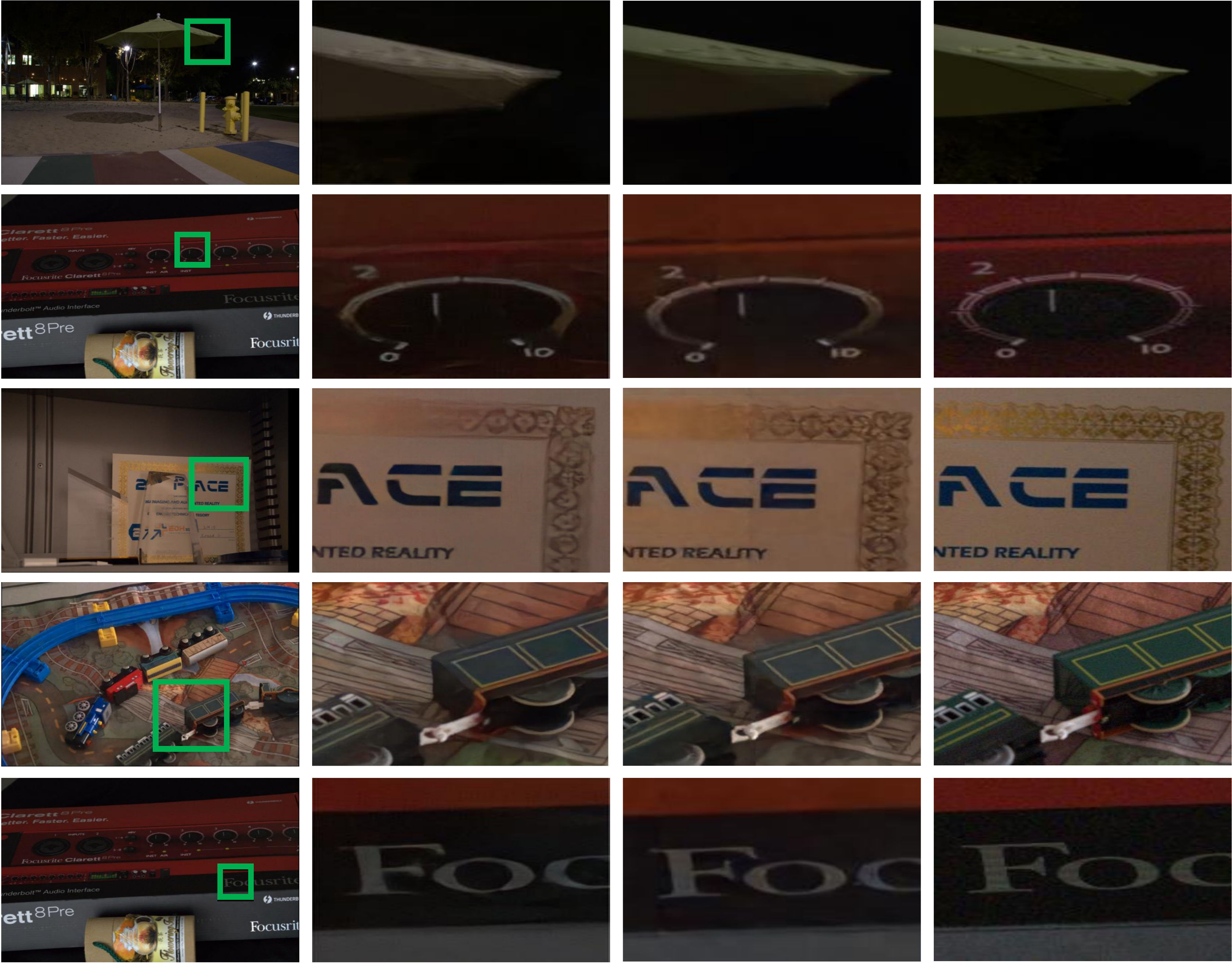}
        \scriptsize HR Image \quad \qquad Karadeniz \etal \cite{karadeniz2020burst} \quad \xnet (Ours) \qquad Ground-truth 
        \vspace{-2mm}
        \caption{Burst low-light image enhancement on Sony subset~\cite{chen2018learning}. \xnet better preserves color and structural details.}
        \label{fig: enhancement}
    \end{figure}

    \vspace{0.2em}
    \noindent \textbf{Enhancement results.} 
    In Table \ref{tab: enhancement}, we report results of several low-light enhancement methods. Our \xnet yields significant performance gain of 3.07 dB over the existing best method~\cite{karadeniz2020burst}. Similarly, the visual examples provided in Fig.~\ref{fig: enhancement} also corroborates the effectiveness of our approach.

    \subsection{Burst Denoising}
            Here, we demonstrate the effectiveness of the proposed \xnet on the burst denoising task. \xnet processes the input noisy sRGB burst and obtains a noise-free image. Since, there is no need to up-sample the extracted features, transpose convolution in the proposed AGU is replaced by a simple group convolution while rest of the network architecture is kept unmodified.
            
            \vspace{0.2em}
            \noindent \textbf{Dataset.} We evaluate our approach on the grayscale and color burst denoising datasets introduced in~\cite{Mildenhall2018BurstDW} and~\cite{Xia2020BasisPN}. These datasets contain 73 and 100 burst images respectively. In both datasets, a burst generated synthetically by applying random translations to the base image. The shifted images are then corrupted by adding heteroscedastic Gaussian noise~\cite{Healey1994RadiometricCC} with variance $\sigma_r^2 + \sigma_s x$. The networks are then evaluated on 4 different noise gains $(\text{1, 2, 4, 8})$, corresponding to noise parameters $(\log(\sigma_r), \log(\sigma_s)) \rightarrow$ $ (\text{-2.2, -2.6})$, $(\text{-1.8, -2.2})$, $(\text{-1.4, -1.8})$, and $(\text{-1.1, -1.5})$, respectively. Note that the noise parameters for the highest noise gain (Gain $\propto$ 8) are unseen during training. Thus, performance on this noise level indicates the generalization of the network to unseen noise. Following~\cite{bhat2021deep1}, we utilized 20k samples from the Open Images~\cite{openimages} training set to generate the synthetic noisy bursts of burst-size 8 and spatial size $128 \times 128$. Our \xnet is trained for 50 epochs both for the grayscale and color burst denoising tasks and evaluated on the benchmark datasets~\cite{Mildenhall2018BurstDW} and~\cite{Xia2020BasisPN} respectively.
            
\begin{figure}[t]
    \centering
        \begin{center}
            \scriptsize
            \includegraphics[width=1\linewidth]{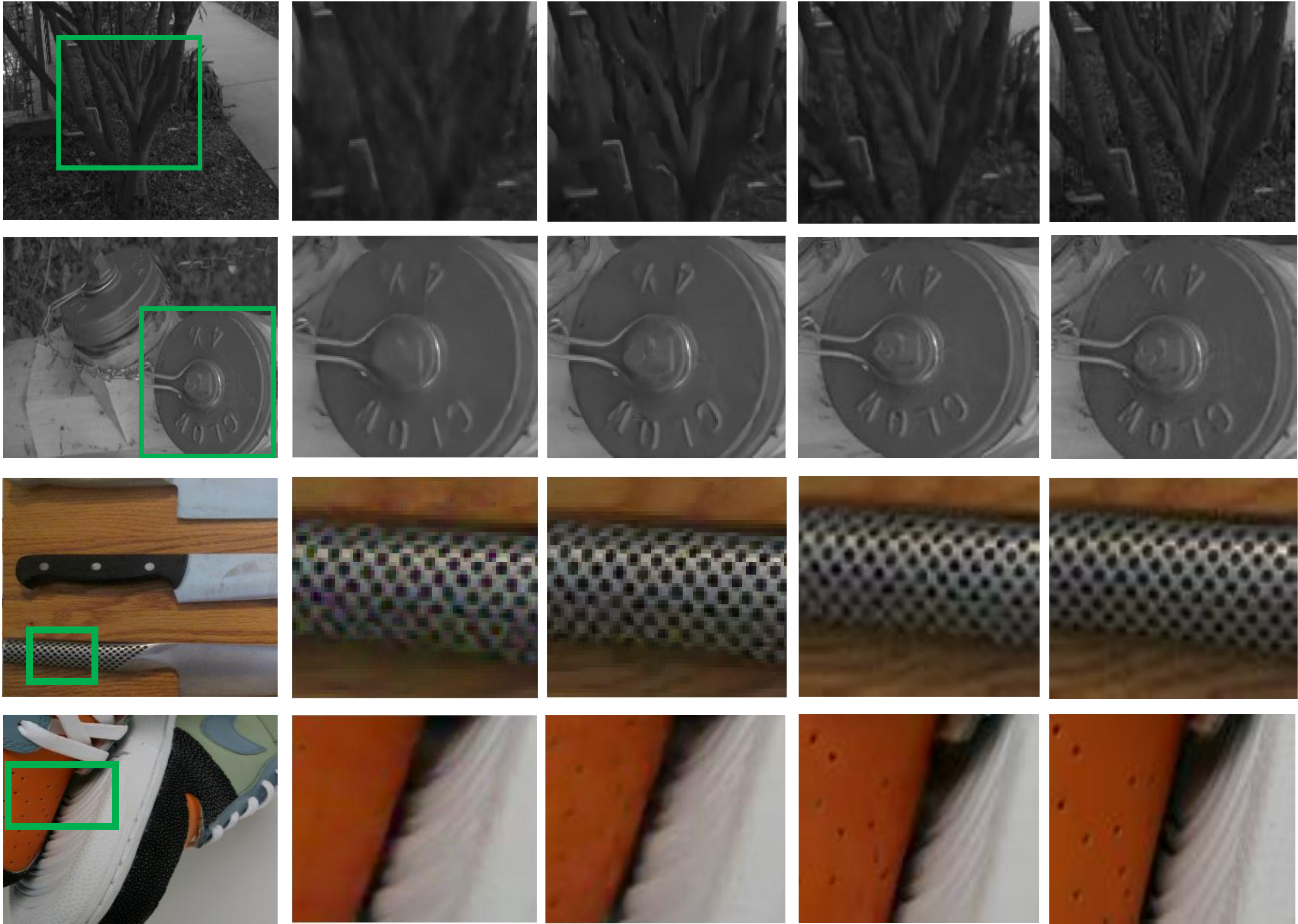}
            \begin{tabular}{rrrrl}
                HR Image & \qquad BPN \cite{Xia2020BasisPN} & \quad MFIR \cite{bhat2021deep1} & \xnet (Ours) & Ground-truth\\
            \end{tabular}\vspace{-6mm}
        \end{center}
    \caption{Comparisons for burst denoising on gray-scale~\cite{Mildenhall2018BurstDW} and color datasets~\cite{Xia2020BasisPN}. Our \xnet produces more sharper and clean results than other competing approaches. Many more examples are provided in the supplementary material.}
    \label{fig: gdn results}
    \end{figure}

            \vspace{0.2em}
            \noindent \textbf{Burst Denoising results.} We compare the proposed \xnet with the several approaches (KPN~\cite{Mildenhall2018BurstDW}, MKPN~\cite{Marinc2019MultiKernelPN}, BPN~\cite{Xia2020BasisPN} and MFIR~\cite{bhat2021deep1}) both for grayscale and color burst denoising tasks. Table~\ref{tab:kpn_grayscale} shows that our \xnet significantly advances state-of-the-art on grayscale burst denoising dataset~\cite{Mildenhall2018BurstDW}. Specifically, the \xnet outperforms the previous best method MFIR~\cite{bhat2021deep1} on all four noise levels. On average, \xnet achieves $2.07$ dB improvement over MFIR~\cite{bhat2021deep1}. Similar performance trend can be observed in Table~\ref{tab:kpn_color} for color denoising on color burst dataset~\cite{Xia2020BasisPN}. Particularly, our \xnet provides PSNR boost of $1.34$ dB over previous best method MFIR~\cite{bhat2021deep1}. Figure~\ref{fig: gdn results} shows that the images reproduced by \xnet are more cleaner and sharper than those of the other approaches.

     \begin{table}[t]
        	\centering\vspace{-1mm}
        	\resizebox{0.99\columnwidth}{!}{%
        		\begin{tabular}{lccccc}
                    \toprule
                    & Gain $\propto$ 1 & Gain $\propto$ 2 & Gain $\propto$ 4 & Gain $\propto$ 8 & Average\\
                    \midrule
                    HDR+~\cite{Hasinoff2016BurstPF}& 31.96  &  28.25 & 24.25  & 20.05&26.13\\
                    BM3D~\cite{Dabov2007ImageDB}& 33.89  &  31.17 & 28.53  & 25.92&29.88\\
                    NLM~\cite{Buades2005ANA}& 33.23  &  30.46 & 27.43  & 23.86&28.75\\
                    VBM4D~\cite{Maggioni2012VideoDD}& 34.60  &  31.89 & 29.20  & 26.52&30.55\\
                    KPN~\cite{Mildenhall2018BurstDW}&36.47&33.93&31.19&27.97&32.39\\
                    MKPN~\cite{Marinc2019MultiKernelPN}&36.88&34.22&31.45&28.52&32.77\\
                    BPN~\cite{Xia2020BasisPN}&38.18&35.42&32.54& 29.45 & 33.90\\
                    MFIR~\cite{bhat2021deep1} & 39.37 & 36.51 & 33.38 & 29.69 & 34.74 \\
                    \midrule
                    \textbf{BIPNet (Ours)} & \textbf{41.26} & \textbf{38.74} & \textbf{35.91} & \textbf{31.35} & \textbf{36.81} \\
                    \bottomrule
                \end{tabular}
        	}\vspace{-2mm}
        	\caption{Comparison of our method with prior approaches on the grayscale burst denoising set~\cite{Mildenhall2018BurstDW} in terms of PSNR. The results for existing methods are from~\cite{bhat2021deep1}.}
        	\label{tab:kpn_grayscale}%
        \end{table}
        
        \begin{table}[t]
        	\centering
        	\resizebox{0.99\columnwidth}{!}{%
        		\begin{tabular}{lccccc}
                    \toprule
                    &Gain $\propto$ 1 & Gain $\propto$ 2 & Gain $\propto$ 4 & Gain $\propto$ 8 & Average \\
                    \midrule
                    
                    KPN~\cite{Mildenhall2018BurstDW}&38.86&35.97&32.79&30.01 & 34.41 \\
                    BPN~\cite{Xia2020BasisPN}&40.16&37.08&33.81&31.19 & 35.56 \\
                    MFIR~\cite{bhat2021deep1} & 42.21 & 39.13 & 35.75 & 32.52 & 37.40 \\
                    \midrule
                    \textbf{BIPNet (Ours)} & \textbf{42.28} & \textbf{40.20} & \textbf{37.85} & \textbf{34.64} & \textbf{38.74} \\
                    \bottomrule
                \end{tabular}
        	}
        	\vspace{-2mm}
        	\caption{Comparison with previous methods on the color burst denoising set~\cite{Xia2020BasisPN} in terms of PSNR. The results for existing methods are from~\cite{bhat2021deep1}. Our approach outperforms BPN on all four noise levels with average margin of 1.34dB.}
        	\label{tab:kpn_color}%
        \end{table}

\section{Conclusion}
    We present a burst image restoration and enhancement framework which is developed to effectively fuse complimentary information from multiple burst frames. Instead of late information fusion approaches that merge cross-frame information towards late in the pipeline, we propose the idea of pseudo-burst sequence that is created by combining the channel-wise features from individual burst frames. To avoid mismatch between pseudo-burst features, we propose an edge-boosting burst alignment module that is robust to camera-scene movements. The pseudo-burst features are enriched using multi-scale information and later progressively fused to create upsampled outputs. Our state-of-the-art results on three burst image restoration and enhancement tasks (super-resolution, low-light enhancement, denoising) corroborate the generality and effectiveness of \xnet. 
    
\section*{Acknowledgement} M.-H. Yang is supported in part by the NSF CAREER Grant 1149783. Authors would like to thank Martin Danelljan, Goutam Bhat (ETH Zurich) and Bruno Lecouat (Inria and DIENS) for their useful feedback and providing burst super-resolution results.

{\small
\bibliographystyle{ieee_fullname}
\bibliography{arxive}
}

\newpage
\renewcommand\thesection{\Alph{section}}
\setcounter{section}{0}

\section*{Supplementary Material}
    Here we describe the architectural details of the proposed \xnet (Sec.~\ref{appendix:network}), 
    and present additional visual comparisons  with existing state-of-the-art approaches for burst SR and burst de-noising (Sec.~\ref{appendix:results1} and Sec.~\ref{appendix:results2}). 

\section{Network Architecture Details}
\label{appendix:network}
    \subsection{Edge Boosting Feature Alignment (EBFA)}
    The proposed feature processing module (FPM) consists of three residual-in-residual (RiR)~\cite{RCAN} groups. Each RiR is made up of three RGCAB and each RGCAB contains a basic residual block followed by a global context attention as shown in Fig. 2 (a) of the main paper. Although, the deformable convolution layer is shown only once in the Fig. 2 (b) for simplicity, we apply three such layers to improve the feature alignment ability of the proposed EBFA module. 

    \subsection{Pseudo Burst Feature Fusion (PBFF)}
    The proposed PBFF is as shown in Fig.~3 (a) in main paper. It consists of multi-scale feature (MSF) extraction module which is made up of a light-weight 3-level U-Net \cite{ronneberger2015unet}. We employed one FPM (with 2 RiR and 2 RGCAB in each RiR) after each downsample and upsample convolution layer. Number of convolution filters are increased by a factor of 1.5 at each downsampling step and decreased by the rate of 1.5 after each upsampling operation. We simply add features extracted at each level to the upsampled features via skip connections.
    
    \subsection{Adaptive Group Up-sampling (AGU)}
    Our AGU module is shown in Fig.~3 (c) in the main paper. It aggregates the input group of pseudo bursts and pass them through a bottleneck convolution layer of kernel size $1$$\times$$1$ followed by a set of four parallel convolution layers, each with kernel size of $1$$\times$$1$ and 64 filters. Further, the outputs from previous step are passed through the softmax activation to obtain the dense attention maps.

\begin{figure*}[t]
    \begin{center}
        \scriptsize
        \begin{tabular}{ccccc}
            \multicolumn{5}{c}{\includegraphics[width=1\textwidth]{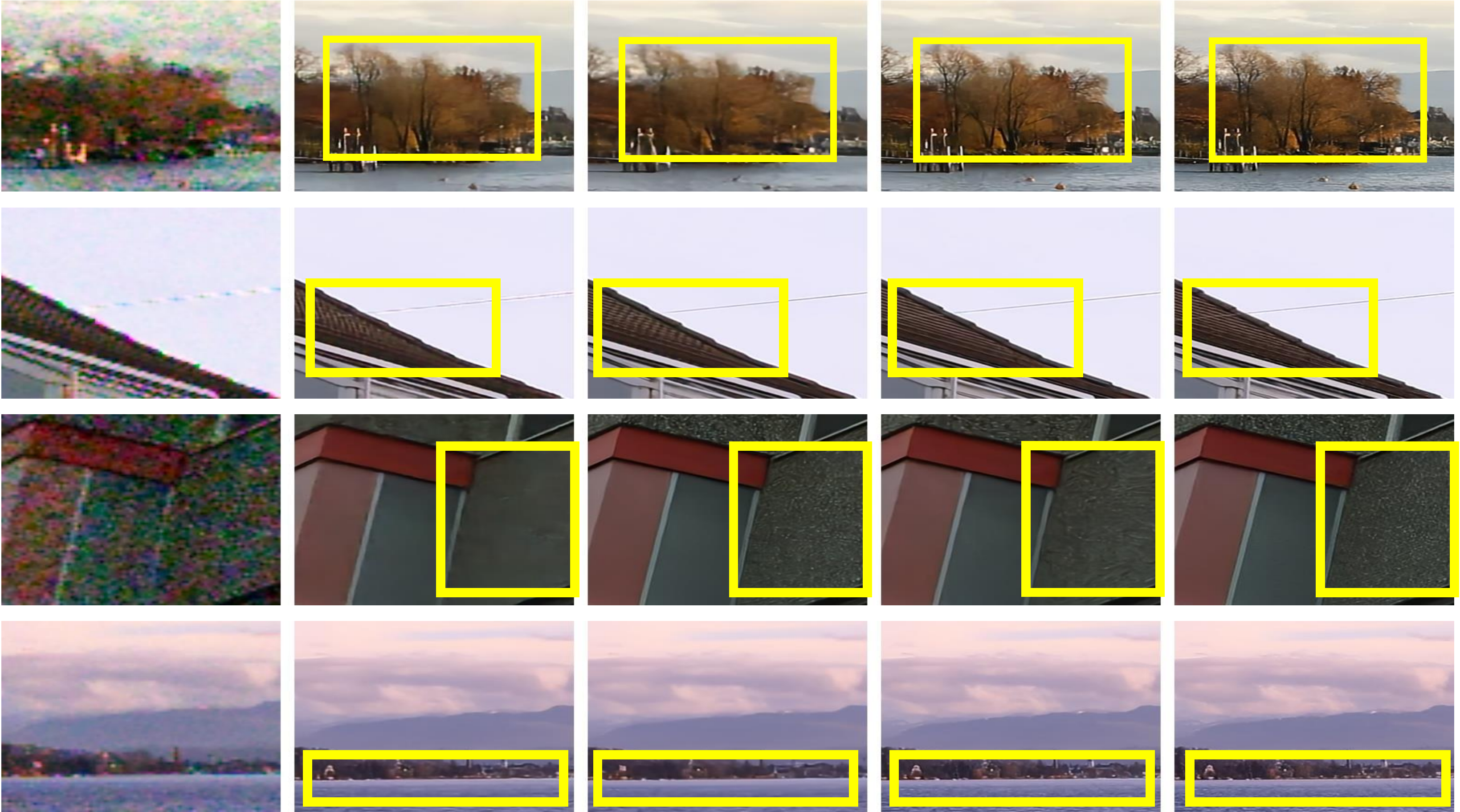}}\\
            \qquad \qquad Base frame & \qquad \qquad \qquad \qquad DBSR~\cite{bhat2021deep} & \qquad \qquad \qquad \qquad LKR~\cite{lecouat2021lucas} & \qquad \qquad \qquad \qquad MFIR~\cite{bhat2021deep1} & \qquad \qquad \xnet (ours)
        \end{tabular}
    \end{center}\vspace{-1.5em}
    \caption{\small Comparisons for $\times4$ burst super-resolution on SyntheticBurst dataset~\cite{bhat2021deep}. Our \xnet produces more sharper and clean results than other competing approaches (specifically the marked green box regions).}
    \label{fig: s1}
\end{figure*}

\section{Additional Visual Results for Burst SR}
\label{appendix:results1}

\begin{figure*}[t]
    \begin{center}
        \scriptsize
        \begin{tabular}{cccccc}
            \multicolumn{6}{c}{\includegraphics[width=1\textwidth]{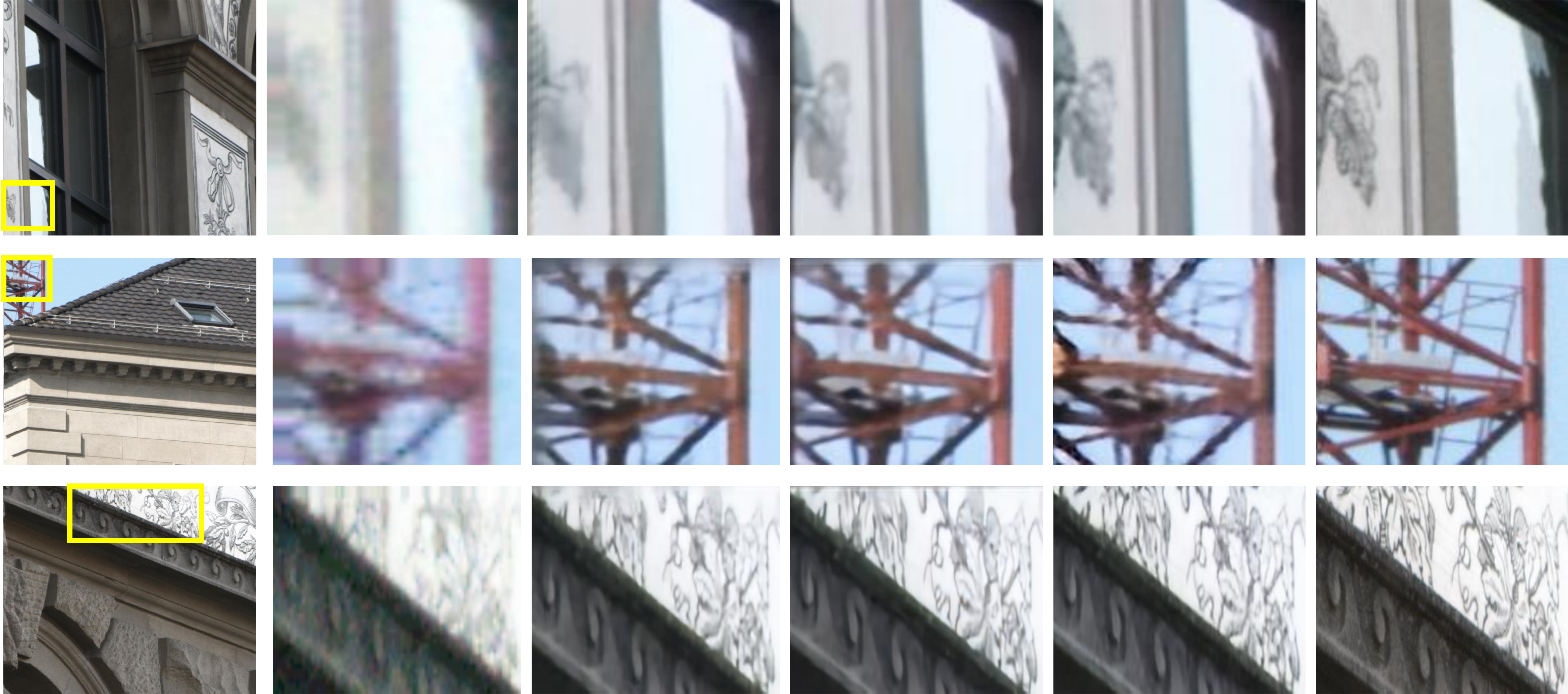}}\\
            \qquad \qquad HR Image & \qquad \qquad \qquad Base frame & \qquad \qquad \qquad DBSR~\cite{bhat2021deep} & \qquad \qquad \qquad MFIR~\cite{bhat2021deep1} & \quad \qquad \qquad \xnet (ours) & \qquad Ground-truth
        \end{tabular}
    \end{center}\vspace{-1.5em}
    \caption{\small Comparison for $\times4$ burst SR on real BurstSR dataset~\cite{bhat2021deep}. 
    The crops shown in red boxes (in the input images shown in the left-most column) are magnified to illustrate the improvements in restoration results.
    The reproductions of our \xnet are perceptually more faithful to the ground-truth than those of other methods.}
    \label{fig: s2}
\end{figure*}

\begin{figure*}[t]
    \begin{center}
        \scriptsize
        \begin{tabular}{ccc}
            \multicolumn{3}{c}{\includegraphics[width=0.9\textwidth]{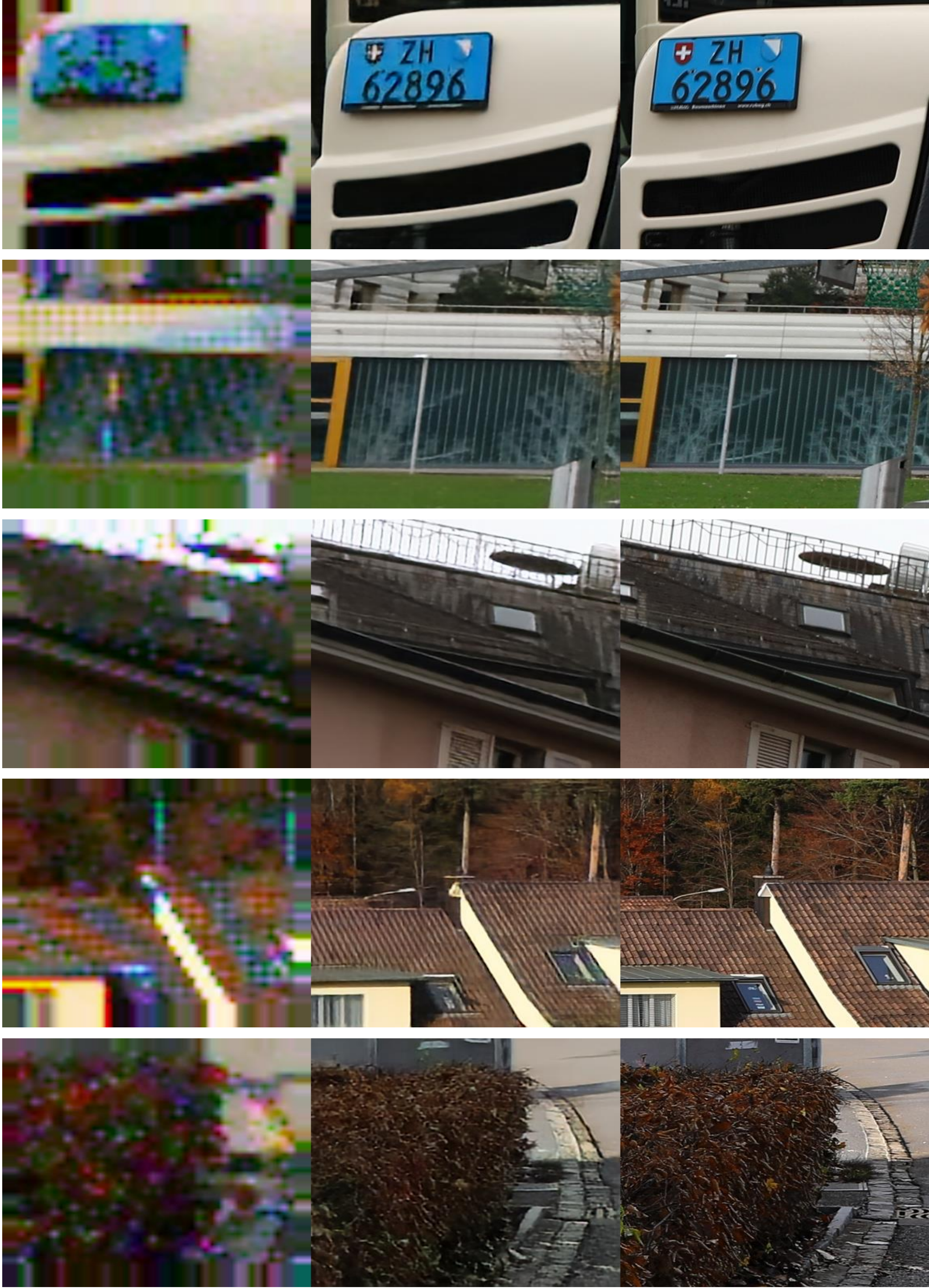}}\\
            \qquad \qquad \qquad \qquad Base frame & \qquad \qquad \qquad \qquad \qquad \qquad \qquad \textbf{BIPNet (Ours)} & \qquad \qquad \qquad \qquad Ground-truth 
        \end{tabular}
    \end{center}\vspace{-1.5em}
    \caption{\small Results for $\times8$ SR on images from SyntheticBurst dataset~\cite{bhat2021deep}. Our method effectively recovers image details in extremely challenging cases.}
    \label{fig: s3}
\end{figure*}

    The results provided in Fig.~\ref{fig: s1} and Fig.~\ref{fig: s2} show that our method performs favorably on both real and synthetic images for the scale factor $\times4$. The true potential of the proposed approach is demonstrated in Fig.~\ref{fig: s3}, where it successfully recovers the fine-grained details from extremely challenging LR burst images (that are down-scaled by a factor of $\times8$).

\section{Additional Results for Burst Denoising}
\label{appendix:results2}

\begin{figure*}[t]
    \centering
        \begin{center}
            \scriptsize
            \includegraphics[width=1\linewidth]{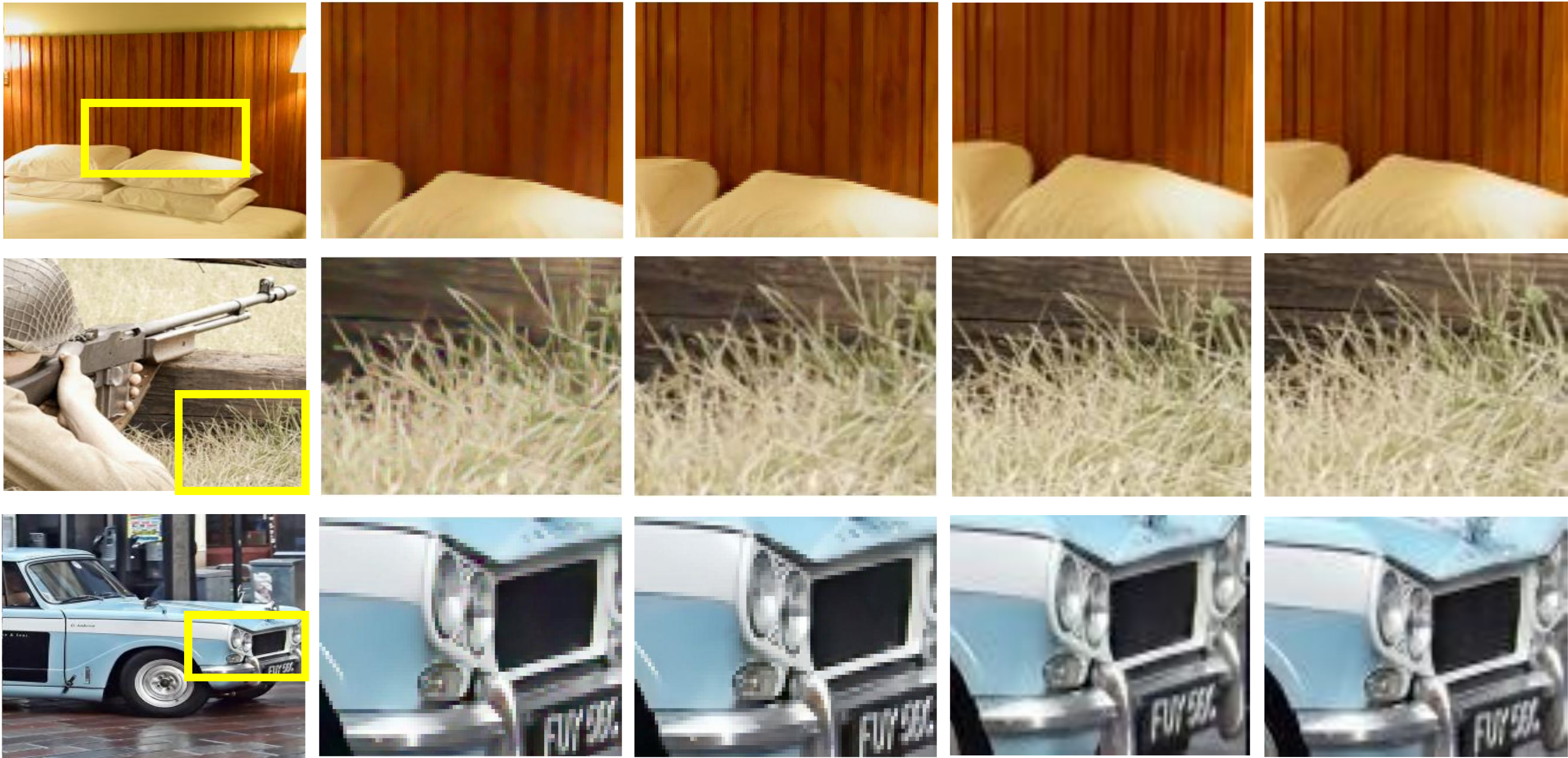}
            \begin{tabular}{p{3cm}p{3cm}p{2cm}p{3cm}p{3cm}}
                HR Image & BPN \cite{Xia2020BasisPN} & MFIR \cite{bhat2021deep1} & \qquad \qquad \xnet (Ours) & \qquad \qquad Ground-truth\\
            \end{tabular}\vspace{-6mm}
        \end{center}
    \caption{\small Comparisons for burst denoising on color datasets~\cite{Xia2020BasisPN}. The crops shown in green boxes (in the input images shown in the left-most column) are magnified to illustrate the improvements in restoration results. Our proposed \xnet produces more sharper and clean results than other competing approaches.}
    \label{fig: s4}
\end{figure*}

\begin{figure*}[t]
    \centering
        \begin{center}
            \scriptsize
            \includegraphics[width=1\linewidth]{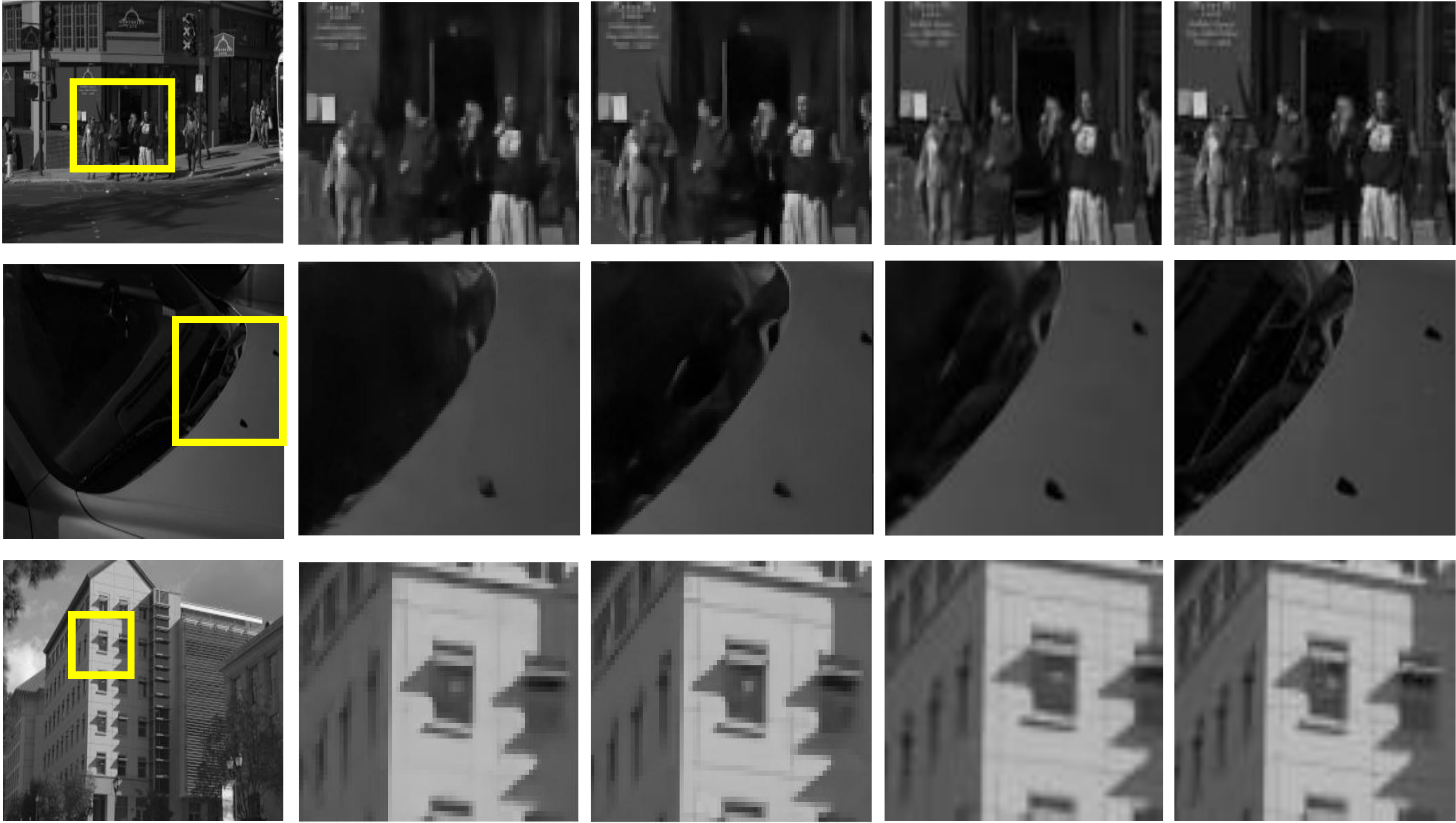}
            \begin{tabular}{p{3cm}p{3cm}p{2cm}p{3cm}p{3cm}}
                HR Image & BPN \cite{Xia2020BasisPN} & MFIR \cite{bhat2021deep1} & \qquad \qquad \xnet (Ours) & \qquad \qquad Ground-truth\\
            \end{tabular}\vspace{-6mm}
        \end{center}
    \caption{\small Comparisons for burst denoising on gray-scale~\cite{Mildenhall2018BurstDW}. The crops shown in green boxes (in the input images shown in the left-most column) are magnified to illustrate the improvements in restoration results. Our \xnet produces more sharper and clean results than other competing approaches.}
    \label{fig: s5}
\end{figure*}

    The results provided in Fig.~\ref{fig: s4} and Fig.~\ref{fig: s5} show that our method performs favorably on both grayscale~\cite{Mildenhall2018BurstDW} and color~\cite{Xia2020BasisPN} noisy images. Specifically, it can recover fine details in the outputs and is more closer to the ground-truth compared to existing state-of-the-art approaches.

\end{document}